\documentclass{article}
\pdfoutput=1
\newcommand*\rot{\rotatebox{90}}
\usepackage{arxiv}
\usepackage{rotating}
\usepackage[utf8]{inputenc} % allow utf-8 input
\usepackage[T1]{fontenc}    % use 8-bit T1 fonts
\usepackage{hyperref}       % hyperlinks
\usepackage{url}            % simple URL typesetting
\usepackage{booktabs}       % professional-quality tables
\usepackage{amsfonts}       % blackboard math symbols
\usepackage{nicefrac}       % compact symbols for 1/2, etc.
\usepackage{microtype}      % microtypography
\usepackage{lipsum}		% Can be removed after putting your text content
\usepackage{graphicx}
\usepackage{doi}

\title{Metric Ensembles for Hallucination Detection}

\author{ 
\hspace{1mm}Grant C. ~Forbes%\thanks{Use footnote for providing further
	\\North Carolina State University\\
    Raleigh, NC \\
	\texttt{gforbes@ncsu.edu} \\
	\And
	\hspace{1mm}Parth~Katlana \\
	North Carolina State University\\
	Raleigh, NC \\
	\texttt{pkatlan@ncsu.edu} \\
	\AND
    \hspace{1mm}Zeydy~Ortiz, Ph. D.\\
    DataCrunch Lab, L. L. C.\\
    Cary, NC\\
    \texttt{zortiz@datacrunchlab.com} \\
}

\date{}

\begin{document}
\maketitle

% Abstract (Do not insert blank lines, i.e. \\) 
\begin{abstract}
Abstractive text summarization has garnered increased interest as of late, in part due to the proliferation of large language models (LLMs). One of the most pressing problems related to generation of abstractive summaries is the need to reduce "hallucinations," information that was not included in the document being summarized, and which may be wholly incorrect. Due to this need, a wide array of metrics estimating consistency with the text being summarized have been proposed. We examine in particular a suite of unsupervised metrics for summary consistency, and measure their correlations with each other and with human evaluation scores in the \textit{wiki\_bio\_gpt3\_hallucination} dataset. We then compare these evaluations to models made from a simple linear ensemble of these metrics. We find that LLM-based methods outperform other unsupervised metrics for hallucination detection. We also find that ensemble methods can improve these scores even further, provided that the metrics in the ensemble have sufficiently similar and uncorrelated error rates. Finally, we present an ensemble method for LLM-based evaluations that we show improves over this previous SOTA.
\end{abstract}

% Keywords
%keyword 1; keyword 2; keyword 3 (List three to ten pertinent keywords specific to the article; yet reasonably common within the subject discipline.)} 

\setcounter{section}{1} %% Remove this when starting to work on the template.

\section{Introduction}

Text summarization is a rapidly changing and advancing field, due in no small part to the advent of Large Language Models (LLMs) such as GPT \cite{bubeck2023sparks} and LaMDA \cite{thoppilan2022lamda}. Many summarization methods, however, struggle with ``hallucinating:" inserting false, misleading and/or nonrepresentative material into the summaries. As such, many automatic methods for hallucination detection have been proposed in the literature for both evaluation and iterative improvement of text summarization methods. This diversity of methods, while indicative of rapid progress, has also led to a situation where there is no one clear standard evaluative metric for hallucinations in text summarization. With this in mind, we test a suite of hallucination detection from prior literature on the \textit{wiki\_bio\_gpt3\_hallucination} dataset \cite{lebret2016neural, manakul2023selfcheckgpt}, and examine their correlations with both each other and with a human evaluation baseline. We also, drawing on prior work in ensemble methods, test these against a linear ensemble of the sampled methods, and found that this ensemble outperforms most individual evaluation metrics. We found evaluation methods based on directly querying LLMs themselves to be most closely correlated with human evaluation, outperforming all non-LLM metrics and the ensemble. With this in mind, we constructed a new ensemble of LLM evaluations with a range of temperatures, with the expectation that perturbations to the metric that didn't correlate with the "true" value of what was being measured would cancel out in aggregate (we elaborate on this expectation in Section~\ref{sec:ensembles}). We found that our LLM ensemble outperformed even the best LLM-based single evaluation, indicating our method to be the most accurate and effective hallucination detection metric to date for our chosen dataset.

\section{Metrics Evaluated and Related Work}

Here, we discuss prior work, and describe the metrics we've chosen to represent the suite of prior methods that exist. We also discuss ensemble methods, the theoretical justifications for their use in this domain, and the conditions generally required for them to be effective.

\subsection{Text Summarization}
 Traditionally, text summarization has been categorized as either extractive or abstractive. Extractive text summarizers, such as OCCAMS \cite{white2023occams}, produce summaries by concatenating particularly salient sentences (``extracts") from the document being summarized. On the other hand, abstractive summarizers, such as most methods using LLMs \cite{zhang2023summit}, attempt to generate a summary "from scratch," assembling new sentences in an attempt to synthesize the information in a document in a more human-seeming way. Abstractive summaries are often able to be more natural-sounding than extractive summaries, but, as has been noted repeatedly in the literature, have a tendency to hallucinate \cite{ji2023survey}. It is often challenging to evaluate these models, as has been noted in \cite{goyal2022news}.

\subsection{Hallucination Detection Metrics}

There are many methods for hallucination detection in prior work: too many to feasibly include them all within this work. We limited the scope of methods that we tested in two key ways: by constraining the broader scope of concern to unsupervised metrics, and by choosing a selection of well-regarded methods meant to cover the breadth of scope within that subfield. When we say we are specifically concerned with unsupervised hallucination detection methods, we mean those which require no input other than the summary and the source text itself. We chose to focus on these metrics as they are the most general, requiring no gold-standard human summaries or other supplementary information, and are thus the most widely deploy-able. More particularly, unsupervised hallucination detection metrics are deploy-able in two important context which exclude any other types of metrics:

1. As an evaluative tool on summarization data (possibly generated continuously, rather than part of a finite set)

2. As an in-the-loop tool for actively curbing hallucinations in a summarization tool at runtime.

The survey \cite{huang2021factual} identifies four general types of unsupervised summarization metrics: ``triple-based," ``textual-entailment-based," ``QA-based," and ``Other." We evaluate representatives from each of these categories (a more detailed analysis of these categories and the intricacies therein can be found in the aforementioned survey). We chose these to both cover the identified breadth of evaluation methods in the literature (i.e., pulling representatives from each of these categories), as well as to find methods with good theoretical/empirical backing and wide use, while still being recently developed and relevant to the SOTA.

\subsubsection{FactSumm}
FactSumm \cite{factsumm} is a triple-based metric to estimate the factual accuracy of generated text. It builds on prior works in graph-based hallucination detection \cite{kryscinski2019evaluating, goodrich2019assessing}, using pre-trained models to extract fact triples (subject, relation, object) from both the source document and the summary, and returns a count of the number of triples extracted from the summary that are included in the extract from the document itself. This serves as a heuristic for the percentage of "facts" in the summary that are contained within the source document.
%The end-to-end models are shown to be able to extract complete
%sets of facts from datasets with full pages of text. It also analyses
%Multiple models that estimate factual accuracy on a Wikipedia text
%summarization task, and show their efficacy compared to ROUGE
%and other model-free variants by conducting a human evaluation
%study.

\subsubsection{QAGS}
QAGS \cite{wang2020asking} is a question-answering based metric that automatically generates questions and answers from the source document and summary, then scores the summary based on how many of the derived questions are answered correctly. We used the implementation of QAGS in \cite{factsumm}. For comparison, other notable representatives from this category are QAEval \cite{deutsch2021towards} and FEQA \cite{durmus2020feqa}.

\subsubsection{ROUGE}
ROUGE \cite{lin2004rouge} is traditionally used as a supervised metric, by calculating the ROUGE score between a generated summary and a gold-standard human summary. However, \cite{10.5555/1699510.1699550} introduced the idea of using ROUGE as an unsupervised metric, and demonstrated its efficacy in such a capacity. Using ROUGE without supervision (``supervision," in this case, referring to gold standard human summaries that can be compared against) involves taking the ROUGE score between the generated summary and the text itself being summarized: treating the text itself, in other words, as its own ``gold standard." The intuition behind this as a heuristic is that hallucinatory passages, on average, are likely to have less similarity (measured by ROUGE) to the source text than those which accurately summarize the source text.

\subsubsection{SMART}
SMART (Sentence Matching for Rating Text) \cite{amplayo2022smart} evaluation works on two principal ideas. Firstly, it treats sentences, rather than tokens, as the basic units of matching between system and reference
summaries. Because, then, exactly matching sentences are most likely nonexistent (in abstractive summaries, though they are trivially present in extractive summaries), SMART utilizes soft-matching functions to compare sentences which can vary with respect to the type of SMART that is being used. SMART types utilized for the purposes of our study are: 
    \begin{itemize}
    \item SMART1: Unigram-based scoring.
    \item SMART2: Bigram-based scoring.
    \item SMARTL: Longest common subsequence-based scoring.
\end{itemize}
    It is also significant to mention that the unit of n-grams used here are chunks of tokens (sentences by
      default). This is different from the token-level n-grams used in standard
      ROUGE.\\
Secondly, SMART allows to compare
the candidate system summary with the source document. This is particularly important when evaluating dimensions of summary quality that rely on the source
document such as factuality.

\subsubsection{SummaC}
SummaC \cite{laban2022summac}, similarly to SMART, runs evaluations on a sentence-by-sentence basis, but unlike SMART, is explicitly based on Natural Language Entailment (NLI) evaluations between sentences in the source document and the summary. SummaC first generates a matrix for every sentence pair between the summary and source document. Then the two models we benchmark analyze this matrix to achieve a benchmark.

SUMMAC$_{zs}$ (Zero-Shot) reduces the pair matrix to a one-dimensional vector by taking the maximum value of each column, then computes the mean. This step retains the score for the document sentence that provides the strongest support for each summary sentence. It leverages the intuition that each sentence in the summary document, if non-hallucinatory, should have at least one sentence in the source document which has a high entailment score.

SUMMAC$_{conv}$ (Convolutional) reduces reliance on extreme values and takes the entire distribution of entailment scores for each summary sentence into account. %The NLI Pair Matrix's columns are converted into fixed-size histograms, representing the distribution of scores for each summary sentence. The histogram is created by binning the NLI scores into evenly spaced bins.
It utilizes a learned convolutional network on the NLI matrix to compute a final score for the respective summary sentence. 

\subsubsection{SelfCheckGPT}

SelfCheckGPT \cite{manakul2023selfcheckgpt} is an unsupervised hallucination detection method that relies on the intuition that factual generated summaries are much more likely to be similar to each other than to those which contain hallucinations, whereas hallucination-containing summaries are not more likely to be similar to each other than to factual summaries. Another way of framing this intuition is that language models which are confident in their knowledge are likely to have much less diverse responses than those which are making things up. It involves generating multiple summaries for a given source document, then utilizing a variety of distance metrics to check generated summaries against each other for similarity, and shows that higher similarity to other generated summaries is highly correlated with human annotations for textual consistency. Note that we specifically benchmarked their unigram-based approach, as it was the single approach that had the highest correlations with human judgements in \cite{manakul2023selfcheckgpt}.

\subsubsection{LLM Self-Evaluation}

Recent work, such as \cite{luo2023chatgpt}, has explored the possibility of using LLMs themselves as evaluative tools for text data generally, and for abstractive summaries in particular. These recent results are very promising, and may usurp traditional, non-LLM-based evaluative methods in this domain, such as those we have listed thus far. For benchmarking these methods, we reproduced the prompting technique described in \cite{luo2023chatgpt}. For ease of reference, the prompt is in Figure~\ref{list:prompt} below.

\begin{figure}
\centering
\hrulefill
%\raggedright 
\begin{verbatim}
Score the following summary given the corresponding article with respect to consistency
from 1 to 10. Note that consistency measures how much information included in the summary
is present in the source article. 10 points indicate the summary contains only statements
that are entailed by the source document. 
Summary: {summary}
Source Article: {source}
Marks:   
\end{verbatim}
\hrulefill
\caption{LLM evaluation prompt}
\label{list:prompt}
\end{figure}

We used this prompt in both GPT 3.5 turbo and GPT 4 models as benchmarks for this method. Our results lend evidence to the claim that these methods indeed surpass more traditional hallucination detection methods, and we use these results to further refine these methods into an ensemble approach that outperforms prior work.

\subsection{Metric Ensembles}\label{sec:ensembles}

It's been long noted that ensembles of models or metrics, even subpar ones combined naively, are surprisingly efficient and can rival or outperform expert judgment \cite{dawes1979robust}. Ensembles also have been noted to aid in explainability in some contexts, something particularly relevant for analysis work, which is very sensitive to reliability and has a high threshold of required trust \cite{forbes2023metric}. Ensemble methods are thus a promising avenue for improving over a baseline, particularly in a domain such as hallucination detection, where there are a wide variety of disparate metrics, none of which is clearly superior to others in measuring the "true" value.

Ensemble methods operate, fundamentally, by leveraging the ability of the individual metrics' error from the ``true" value to cancel each other out in aggregate. As derived in \cite{perrone1995networks}, if there is a collection of value-estimating functions $f_i$, each of which differs from some true function $f$ by some $m_i = f-f_i$, and we assume the errors are uncorrelated, then in expectation we should expect the error $m_{sum}$ of $f_{sum}$ to be
\begin{equation}
    m_{sum} = f - \frac{1}{N} \sum_{i = 0}^{N-1} f_i = \frac{\bar{m_i}}{N}
\end{equation}
where $\bar{m_i}$ is the mean value of $M_i$. Due to this minimization of error by a factor of $N$, ensembles are a powerful tool to deploy in spaces, such as hallucination detection, where we have many diverse estimates for ground truth, but no (or prohibitively slow and expensive) access to that ground truth itself. There are then two qualities of some collection of metrics $f_i$ that we would want, in order for an ensemble method to be effective:

\textbf{Condition 1.} The metrics must be diverse: that is, their errors should be relatively uncorrelated with each other (this assumption is key for the referenced derivation in \cite{perrone1995networks}).

\textbf{Condition 2.} The metrics must have $m_i$'s that are similar in magnitude. If this condition is not met, then it is possible that the $f_i$ with the lowest error alone would outperform an ensemble model. In other words, the ensemble should only be derived from models that are similarly good estimators of the true function $f$.

The metrics we're using in this work certainly meet condition 1, as we've selected them to cover the breadth of methods in the literature. It remains to be seen, however, if they meet condition 2, and in fact we shall see that, as selected, they do not yet meet this condition. 

Some recent prior work has used ensemble, or ensemble-esque methods to leverage LLMs effectively. These often involve iterative prompting techniques, such as in \cite{du2023improving}, which prompts agents to "debate" each other before arriving at a final answer. SelfCheckGPT \cite{manakul2023selfcheckgpt} itself could be seen as a variation of an ensemble method, as it involves self-checking the model's responses against other responses it might have given. Similar work is in \cite{wang2022self}, which incorporates a self-checking ensemble approach into the sampling algorithm for an LLM.

While in this work, we simply use naive similar ensembles of metrics with uniform weights, as has been shown to be effective \cite{dawes1979robust, perrone1995networks}, some other work has been done on using unlabelled data to find the best term weights for metrics \cite{platanios2014estimating,platanios2016estimating, platanios2017estimating}. While this particular line of work is applied specifically to binary classifiers, and we're working with metrics that cannot be generally constricted to outputs $\in \{0,1\}$, we believe something like this approach could be extended to metric ensembles in future work.

\section{Method}\label{method}

We used the \textit{wiki\_bio\_gpt3\_hallucination}  dataset. This dataset consists of a subset of 238 entries from the original Wikibio dataset, generated in \cite{lebret2016neural}, accompanied by GPT3-generated summaries and sentence-by-sentence human evaluations of those summaries, ranking each sentence as "accurate," "minor inaccurate," or "major inaccurate." These additional summaries and human evaluations were generated in \cite{manakul2023selfcheckgpt}. For each summary in the dataset, we evaluate each metric on this summary and the source document from which it was generated. We then compute the Pearson correlation between this metric and the other metrics, including human evaluation, which we treat as our ``gold standard" metric, or ground truth. For this ground truth, we translate the human evaluations into a single scalar value by taking their mean, wherein we treat ``major inaccurate" as a 0, ``accurate" as a 1, and ``minor inaccurate" as a 0.5. Note that this is the reverse of the method used in \cite{manakul2023selfcheckgpt}: we chose this in order to align the direction of our gold standard with our other hallucination detection benchmarks, in which higher numbers consistently indicate good summaries, rather than bad.

For our ensemble, we simply took a weighted mean of the other models. Weights were set to a constant of 1 for each model, except for in cases where two models were simply closely-aligned permutations of the same model, in which case they were treated as ``sub-models" of the same base model, which would not have the required diversity to meet Condition 1 in Section~\ref{sec:ensembles}. In these cases, each sub-model was weighted by some fraction chosen according to how many sub-models were being used, so that they sum to 1. Going down the rows of Table \ref{tableall} (excluding the first and last rows), the weights used were $[1, 1, \frac{1}{3}, \frac{1}{3}, \frac{1}{3}, \frac{1}{3}, \frac{1}{3}, \frac{1}{3}, \frac{1}{2}, \frac{1}{2}, 1, \frac{1}{2}, \frac{1}{2}]$.

\section{Results}

Our results are split into roughly two sections: those comparing all metrics or ensembles of those metrics, and those which focus exclusively on "Non-LLM" metrics. Note that this term, as we're using it, refers to all metrics that do not involve a direct evaluation from an LLM: so SelfCheckGPT \cite{manakul2023selfcheckgpt}, while ostensibly running functions over many LLM outputs, is evaluated in only the former of these sections, and not the latter. Section \ref{nonllm} covers all metrics, while Section \ref{llm} covers the further dedicated experiments on solely LLM-based metrics/ensembles.

\subsection{Non-LLM Metric Correlations}\label{nonllm}

\begin{figure}[htbp]
  \centering
  \includegraphics[width=0.6\textwidth]{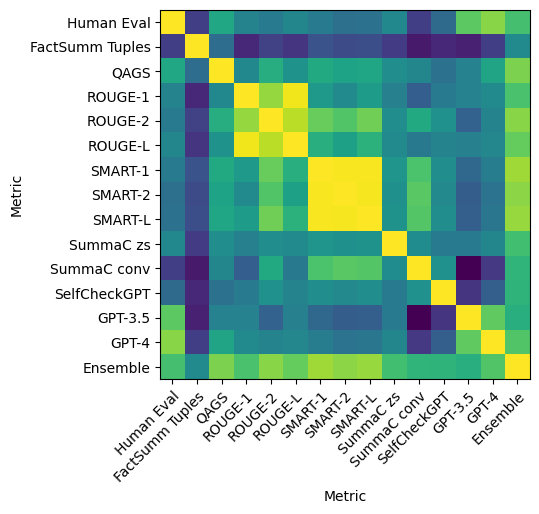}
  \caption{Heatmap of Pearson correlations between all benchmark metrics, our linear ensemble of all benchmarks, and human evaluations.}
  \label{heatmapall}
\end{figure}

\begin{figure}[htbp]
  \centering
  \includegraphics[width=0.6\textwidth]{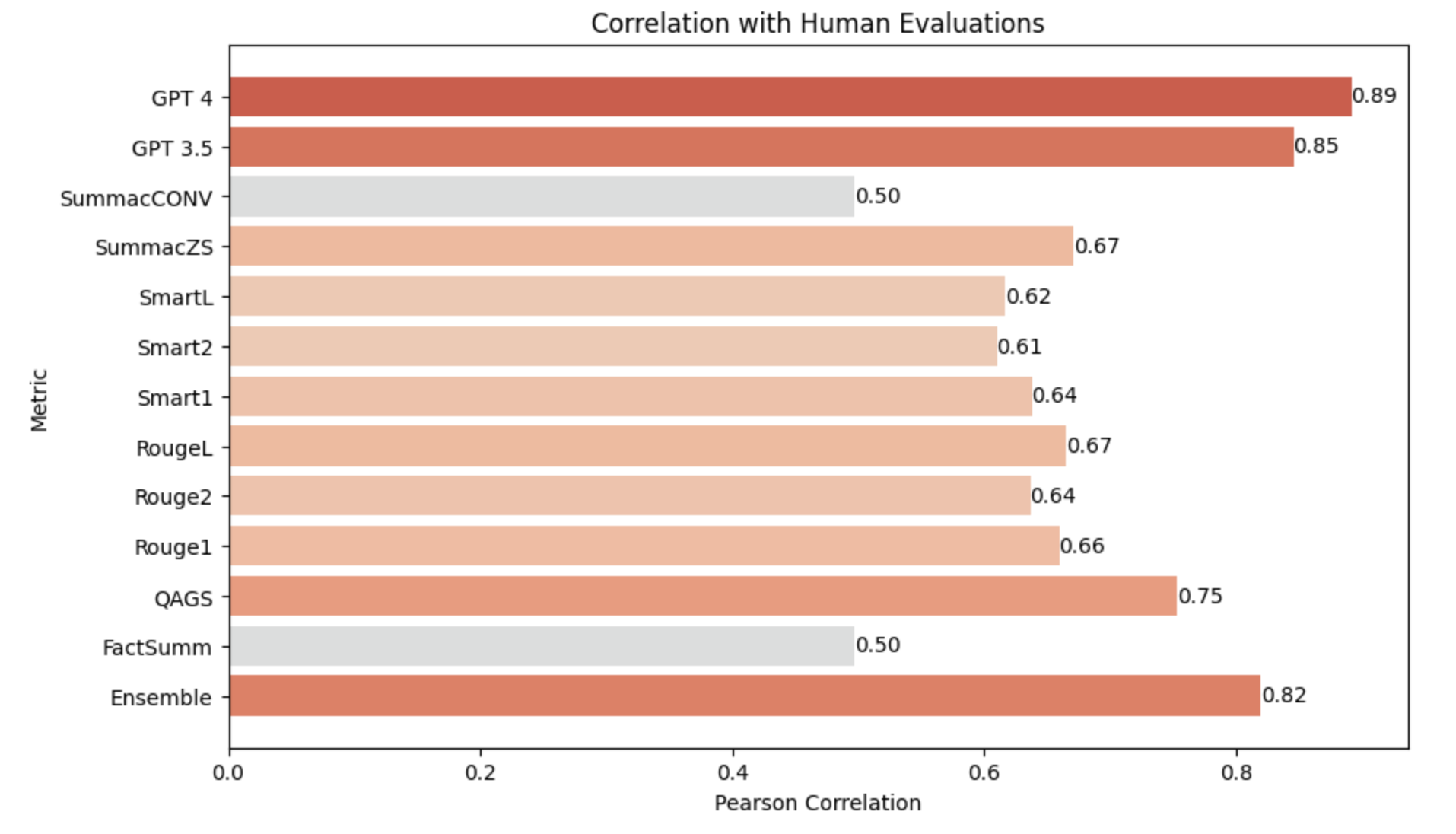}
  \caption{Plot of Pearson correlation with human judgment for each model.}
  \label{barall}
\end{figure}

The correlations between all benchmarks, human evaluations, and our linear ensemble method are recorded in Table \ref{tableall}, and displayed visually as a heatmap in Figure \ref{heatmapall}.
Additionally, a plot showing just the correlations of each method with human evaluations (corresponding to the topmost/leftmost row/column in Figure \ref{heatmapall}) is shown in Figure \ref{barall}.

\begin{table} 
\centering  
\resizebox{\textwidth}{!}{
    \begin{tabular}{@{} cl*{15}c @{}}
  
        & & \rot{Human Eval} & \rot{FactSumm Tuples} & \rot{QAGS} & \rot{ROUGE-1} 
        & \rot{ROUGE-2} & \rot{ROUGE-L} & \rot{SMART-1} 
        & \rot{SMART-2} & \rot{SMART-L} & \rot{SummaC$_{zs}$}& \rot{SummaC$_{conv}$} & \rot{SelfCheckGPT} & \rot{GPT-3.5} & \rot{GPT-4} &\rot{Ensemble}\\
        \cmidrule{2-17}
        & Human Eval             & 1.00 & 0.50 &0.75& 0.66& 0.64& 0.67&
  0.64& 0.61& 0.62 &0.67& 0.50& 0.60&
  0.85 &0.89 &0.82 \\
        & FactSumm Tuples              & 0.50& 1.00 & 0.60 & 0.45 & 0.50 & 0.47& 0.54 &0.52& 0.53 
        & 0.49 & 0.43 & 0.46& 0.44& 0.50 &0.68\\
        & QAGS &0.75 & 0.60& 1.00 & 0.67 & 0.77& 0.70 &0.76&0.74 &0.75 & 0.69 &0.67 &0.61 &0.66 &0.74 &0.89\\
        & ROUGE-1 &0.66 & 0.45 & 0.67 & 1.00 & 0.90 & 0.99 & 0.72 & 0.68 & 0.72 & 0.65 & 0.57 & 0.63 & 0.66 & 0.68 & 0.82\\
        & ROUGE-2    &0.64 & 0.50 & 0.77 & 0.90 & 1.00 & 0.94 & 0.86 & 0.83 & 0.87 & 0.68 & 0.76 & 0.69 & 0.58 & 0.66 & 0.89\\            
        & ROUGE-L    &0.67 & 0.47 & 0.70 & 0.99 & 0.94 & 1.00 & 0.77 & 0.73 & 0.78 & 0.67 & 0.64 & 0.66 & 0.65 & 0.67 & 0.85\\           
      & SMART-1    &0.64 & 0.54 & 0.76 & 0.72 & 0.86 & 0.77 & 1.00 & 0.99 & 0.99 & 0.71 & 0.83 & 0.69 & 0.59 & 0.64 & 0.91\\            
      & SMART-2     & 0.61 & 0.52 & 0.74 & 0.68 & 0.83 & 0.73 & 0.99 & 1.00 & 0.99 & 0.70 & 0.84 & 0.68 & 0.56 & 0.62 & 0.89\\           
%       \rot{\rlap{~Metrics}}
      & SMART-L            & 0.62 & 0.53 & 0.75 & 0.72 & 0.87 & 0.78 & 0.99 & 0.99 & 1.00 & 0.70 & 0.83 & 0.68 & 0.57 & 0.63 & 0.91\\    
      & SummaC$_{zs}$       &0.68 & 0.49 & 0.69 & 0.65 & 0.68 & 0.67 & 0.71 & 0.69 & 0.70 & 1.00 & 0.68 & 0.64 & 0.64 & 0.67 & 0.81        \\         
      & SummaC$_{conv}$     &0.50 & 0.43 & 0.67 & 0.57 & 0.76 & 0.64 & 0.83 & 0.84 & 0.83 & 0.68 & 1.00 & 0.70 & 0.39 & 0.49 & 0.79 \\  
      & SelfCheckGPT        & 0.60 & 0.46 & 0.61 & 0.63 & 0.69 & 0.66 & 0.69 & 0.67 & 0.68 & 0.64 & 0.70 & 1.00 & 0.48 & 0.57 & 0.78\\  
      & GPT-3.5             &0.85 & 0.44 & 0.66 & 0.66 & 0.58 & 0.65 & 0.59 & 0.56 & 0.57 & 0.64 & 0.39 & 0.48 & 1.00 & 0.85 & 0.77 \\
      & GPT-4               & 0.89 & 0.50 & 0.74 & 0.68 & 0.66 & 0.67 & 0.64 & 0.62 & 0.63 & 0.67 & 0.49 & 0.57 & 0.85 & 1.00 & 0.83\\
      & Ensemble               &0.82 & 0.68 & 0.88 & 0.82 & 0.89 & 0.85 & 0.91 & 0.89 & 0.91 & 0.81 & 0.79 & 0.78 & 0.77 & 0.83 & 1.00\\

        \cmidrule[1pt]{2-17}
        
    \end{tabular}
    }
   
    \caption{Pearson correlations between all metrics, our linear ensemble, and human evaluations in the WikiBio hallucination dataset. }
    \label{tableall}
\end{table}

We note and discuss several of the more salient data from this study.

\subsubsection{Our Ensemble Outperforms All Non-LLM-Based Methods}

We observe that the LLM-based methods have the highest correlation with human evaluations, but that the ensemble method outperforms all others. \footnote{For further context, if the LLM-based models are removed from the ensemble, it outperforms all remaining methods other than QAGS, and if both the LLM-based and QAGS metrics are removed from the ensemble, it outperforms all remaining methods.} The fact that some of the base methods used in the ensemble outperform the ensemble itself suggests that, in this particular case, Condition 2 of Section 3.3 is not satisfied: there are certain methods (GPT-3 and GPT-4, used as in \cite{luo2023chatgpt}) that outperform the others to a significant enough extent that the gains from the ensemble's diversity are not sufficient to overcome the losses from giving more weight to less accurate models. It is for this reason that we conduct additional experiments in Section \ref{llm}, focusing exclusively on the LLM-based methods.

\subsubsection{Methodological Similarities Yield High Correlations}  

As we would expect, blocks of methods that are methodologically similar yield high Pearson correlations with each other. Particularly, the ROUGE-based methods are all highly correlated with each other, as are the SMART-based methods. These two blocks are also somewhat highly correlated with each other, likely due to the fact that they're doing very similar things under the hood (though SMART is doing this with sentences as the basic blocks of analysis, whereas ROUGE is doing it with tokens).

SummaC and LLM-based methods, alternatively, while slightly higher-correlated with methodologically similar methods, are not as homogeneous as these. 

\subsubsection{Comparative Performance Measured By Human Evaluation}

As noted previously, LLM-based methods and QA-based methods both perform well on this dataset. Our graph-based method performed surprisingly poorly: without further study, however, it cannot be said if this is an indictment of graph-based hallucination detection methods as a whole, or simply of the particular tuple-generating models employed by FactSumm. It is possible that a more complex or robust graph-generating model would considerably boost performance here.

The zero-shot version of SummaC outperformed the convolutional version by a substantial margin, which is surprising, as it goes against the findings in the original SummaC paper. We take this as suggesting that the convolutional model weights might be overfit to the CNN dailymail dataset that the model was trained on, and that a more ``commonsense" zero-shot model may actually be more generalizable.

%\documentclass{article}
%\usepackage{booktabs}

%\begin{table}
%\centering
%\begin{tabular}{cl*{4}{c}}
%\toprule
%& & \multicolumn{4}{c}{Metrics} \\[2ex]
%\cmidrule{3-6}
%& & ROUGE-1 & ROUGE-2 & ROUGE-L & SummacCONV \\
%\cmidrule{2-6}
%& SMART-1  & 0.72 & 0.86 & 0.77 & 0.85 \\
%& SMART-2 & 1.00 & 0.50 & 0.75 & 0.80 \\
%& SMART-l & 1.00 & 0.50 & 0.75 & 0.90 \\
%\cmidrule[1pt]{2-6}
%\end{tabular}
%\caption{Pearson Correlation}
%\end{table}

%\textbf{SummacCONV and SMART Correlation:}
%SummacCONV exhibits some correlation with SMART, although it is not as strong as the Rouge-SMART correlation. This suggests that SummacCONV shares some common ground with SMART in terms of evaluating NLP tasks but also has distinct characteristics that contribute to the moderate correlation.

%\textbf{Low Correlation of Other Metrics:}
%On the other hand, many other metrics do not perform as well and demonstrate low correlation values. This implies that these metrics may measure different aspects of NLP evaluation compared to Rouge, SMART, and SummacCONV.

%Given the low correlation of several metrics and the moderate correlation between SummacCONV and SMART, there is a clear indication for the need of an ensemble approach. An ensemble method can be utilized to combine the strengths of multiple metrics and improve the overall evaluation performance for NLP tasks. This will help in obtaining a more comprehensive and robust assessment of the models or systems under evaluation.

\subsection{LLM Metric Correlations}\label{llm}
\begin{figure}[htbp]
  \centering
  \includegraphics[width=0.6\textwidth]{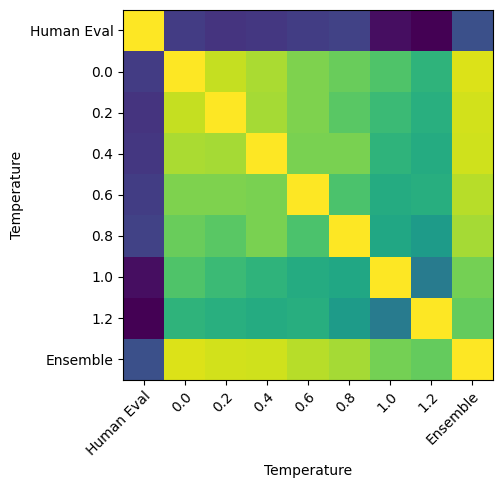}
  \caption{Heatmap of Pearson correlations between GPT-4 evaluations at various temperatures, our linear GPT-4 ensemble, and human evaluations.}
  \label{heatmapgpt}
\end{figure}

\begin{figure}[htbp]
  \centering
  \includegraphics[width=0.6\textwidth]{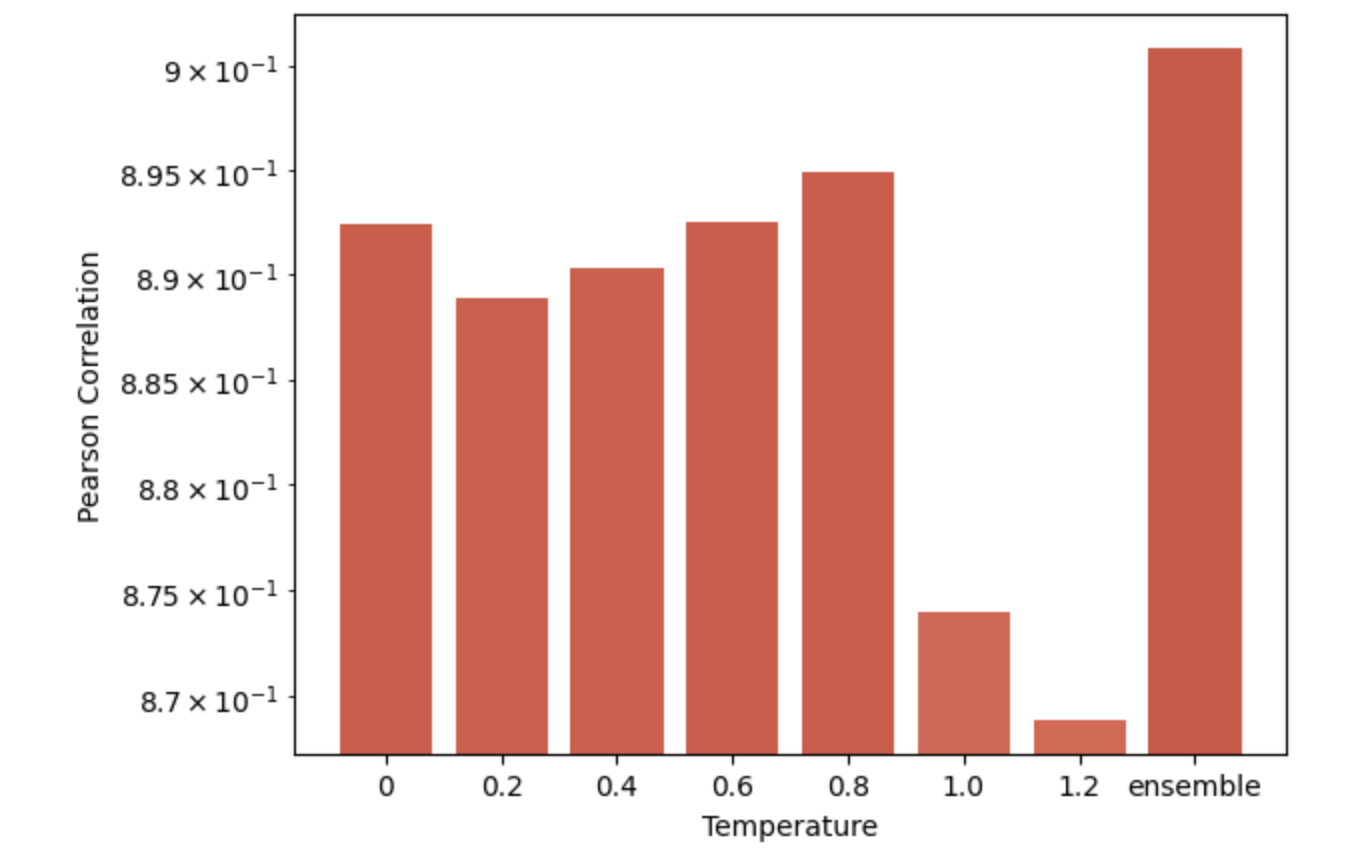}
  \caption{Plot of Pearson correlation with human judgement for each model. Note the log axis here, used to better visualize differences, as models in this ensemble had consistently higher correlations, and were more closely clustered together.}
  \label{bargpt}
\end{figure}

We ran the prompt in Figure \ref{list:prompt} in GPT-4, at different temperature settings, ranging from 0 to 1.2. We also made an ensemble of these models' predictions, using the ensembling method described in Section \ref{method}\footnote{Here, unlike in Section \ref{nonllm}, the weights for each model in the ensemble were set to 1, as there was no structure to the underlying models' similarity to each other that would suggest a need for 'grouping.'}. We then compared the Pearson correlations between each model and the other models, as well as the model ensemble and human evaluations. Our results are collected in Table \ref{temptable}, and a heatmap is depicted in Figure \ref{heatmapgpt}. A bar plot specifically showing the correlations with human judgement is shown in Figure \ref{bargpt}. 

We will now discuss some of the more salient aspects of these results.

%\iffalse

\begin{table} \centering
  \resizebox{\textwidth}{!}{
    \begin{tabular}{@{} cl*{9}c @{}}
        & & Human Eval & 0 & 0.2 & 0.4 
        & 0.6 & 0.8 & 1.0 
        & 1.2 & Ensemble \\
        \cmidrule{2-11}
        & Human Eval       & 1.000 & 0.892 & 0.889 & 0.890 & 0.893 & 0.895 & 0.874 & 0.869 & 0.901 \\
        & 0                & 0.892 & 1.000 & 0.988 & 0.984 & 0.975 & 0.970 & 0.964 & 0.954 & 0.993  \\
        & 0.2              & 0.889 & 0.988 & 1.000 & 0.983 & 0.975 & 0.966 & 0.958 & 0.952 & 0.991 \\
        & 0.4              & 0.890 & 0.984 & 0.983 & 1.000 & 0.973 & 0.974 & 0.954 & 0.950 & 0.990  \\
        & 0.6              & 0.893 & 0.975 & 0.975 & 0.973 & 1.000 & 0.963 & 0.950 & 0.951 & 0.986  \\
        & 0.8              & 0.895 & 0.970 & 0.966 & 0.974 & 0.963 & 1.000 & 0.947 & 0.941 & 0.982 \\
        & 1.0              & 0.874 & 0.964 & 0.958 & 0.954 & 0.950 & 0.947 & 1.000 & 0.923 & 0.973 \\
        & 1.2              & 0.869 & 0.954 & 0.952 & 0.950 & 0.951 & 0.941 & 0.923 & 1.000 & 0.969 \\
        & Ensemble         & 0.901 & 0.993 & 0.991 & 0.990 & 0.986 & 0.982 & 0.973 & 0.969 & 1.000  \\

        \cmidrule[1pt]{2-11}
    \end{tabular}
    }
    \caption{Pearson correlation of different temperatures with human evaluations and the LLM ensemble. }
    \label{temptable}
\end{table}

\subsubsection{Our LLM Ensemble Outperforms All Other Methods}

Most significantly, we find that our naive constant-weighted sum of GPT-4 temperatures outperformed every temperature in isolation. This suggests that the conditions of Section \ref{sec:ensembles} are met in this domain, when we focus specifically on models that are permutations of the LLM-based method in \cite{luo2023chatgpt}. These SOTA results suggests that LLM ensemble methods may be a general tool to improve over a baseline of LLM performance, at least in domains such as this wherein the output being requested of the LLM is a single number, and thus they can be combined straightforwardly in a linear ensemble\footnote{It remains to be seen the extent to which LLM-ensembling can improve performance in more complex domains; this is likely highly dependent on the effectiveness of how the disparate LLM outputs are 'combined.'}. 

\subsubsection{Diversity Increases With Temperature}

As one would expect, higher-temperature GPT-4 runs are less correlated with each other than lower-temperature runs. Perhaps less intuitively, the higher-temperature runs are individually less correlated with each other than either is with the lower temperature runs: i.e., 0.6 and 0.8 are less correlated with each other than either is with 0.0, despite being technically more methodologically similar to each other in a naive sense. This suggests that the 0-temperature run acts as a sort of 'centroid' around which the other runs are clustered: though notably, it is not as 'central' (and, importantly, not as correlated with human evaluations) as the temperature ensemble.

\subsubsection{Lower Temperatures Are Not Straightforwardly Associated With Better Results}

While both previous literature and prevailing wisdom has suggested that LLMs tend to perform better at lower temperatures, our results suggest a perhaps more nuanced picture. Our highest-performing non-ensemble temperature was 0.8, suggesting that some degree of stochasticity can improve overall performance. Notably, however, there is, as one would expect, such a thing as too much stochasticity, as the results decline rapidly from there.

\section{Discussion and Conclusion}

We've presented a review and appraisal of a representative sample of hallucination detection methods from recent literature, and found that LLM-based methods substantially outperform more traditional methods, in terms of their correlation with human evaluations on the Wikibio-gpt3 hallucination detection dataset. We've also tested an ensemble of LLM-based methods across temperature settings, and found this to outperform baseline LLM-based methods, suggesting a new SOTA for improving on LLM-based evaluation. We suggest that this indicates a new standard for LLM-based evaluation in which the LLMs are queried multiple times, and their answers combined, as a way to consistently improve performance over baseline LLM-based evaluation methods. In future work, it would be beneficial to extend this analysis both to more complex ensembling methods, such as those in \cite{platanios2014estimating,platanios2016estimating,platanios2017estimating}, as well as to more complex LLM domains, in which the outputs of the LLM are not numerical values, and thus perhaps more difficult to straightforwardly combine. Some contemporary work, such as \cite{du2023improving, wang2022self, manakul2023selfcheckgpt}, can be viewed partially as an extension of an ensembling approach to these more complex domains, and thus as evidence that this is a promising future direction for research. We hope the evaluation of hallucination detection methods we've presented here can help both to lend evidence to this more complex and integrated use of LLMs, as well as to elucidate the state of black box hallucination detection methods, and their relative strengths and drawbacks relative to each other.%Abstractive summaries are prone to hallucinations, meaning they may include statements that lack support from the original text. Some of these statements can be outright false, while others may be unsupported due to insufficient evidence within the source document. To address this issue, prior research has introduced several fact-checking tools that rely on automatic question-answering systems and textual entailment methods.\cite{louis2022opinesum,10.5555/1699510.1699550}

\newpage
\bibliography{template}
\bibliographystyle{alpha}

% If authors have biography, please use the format below
%\section*{Short Biography of Authors}
%\bio
%{\raisebox{-0.35cm}{\includegraphics[width=3.5cm,height=5.3cm,clip,keepaspectratio]{Definitions/author1.pdf}}}
%{\textbf{Firstname Lastname} Biography of first author}
%
%\bio
%{\raisebox{-0.35cm}{\includegraphics[width=3.5cm,height=5.3cm,clip,keepaspectratio]{Definitions/author2.jpg}}}
%{\textbf{Firstname Lastname} Biography of second author}

% For the MDPI journals use author-date citation, please follow the formatting guidelines on http://www.mdpi.com/authors/references
% To cite two works by the same author: \citeauthor{ref-journal-1a} (\citeyear{ref-journal-1a}, \citeyear{ref-journal-1b}). This produces: Whittaker (1967, 1975)
% To cite two works by the same author with specific pages: \citeauthor{ref-journal-3a} (\citeyear{ref-journal-3a}, p. 328; \citeyear{ref-journal-3b}, p.475). This produces: Wong (1999, p. 328; 2000, p. 475)

%%%%%%%%%%%%%%%%%%%%%%%%%%%%%%%%%%%%%%%%%%
%% for journal Sci
%\reviewreports{\\
%Reviewer 1 comments and authors’ response\\
%Reviewer 2 comments and authors’ response\\
%Reviewer 3 comments and authors’ response
%}
%%%%%%%%%%%%%%%%%%%%%%%%%%%%%%%%%%%%%%%%%%
%\PublishersNote{}
%\end{adjustwidth}
\end{document}